# Probabilistic Wildfire Spread Prediction Using an Autoregressive Conditional Generative Adversarial Network

**Taehoon Kang[1], and Taeyong Kim[2*]**

[1] Department of Civil Systems Engineering, Ajou University, Suwon, Republic of Korea
[2] Department of Civil Systems Engineering, Ajou University, Suwon, Republic of Korea
* Correspondence: taeyongkim@ajou.ac.kr; Tel.: +82-31-219-2505

**Abstract:** Climate change has intensified the frequency and severity of wildfires, making rapid and accurate prediction of fire spread essential for effective mitigation and response. Physics-based simulators such as FARSITE offer high-fidelity predictions but are computationally intensive, limiting their applicability in real-time decision-making, while existing deep learning models often yield overly smooth predictions that fail to capture the complex, nonlinear dynamics of wildfire propagation. This study proposes an autoregressive conditional generative adversarial network (CGAN) for probabilistic wildfire spread prediction. By formulating the prediction task as an autoregressive problem, the model learns sequential state transitions, ensuring long-term prediction stability. Experimental results demonstrate that the proposed CGAN-based model outperforms conventional deep learning models in both overall predictive accuracy and boundary delineation of fire perimeters. These results demonstrate that adversarial learning allows the model to capture the strong nonlinearity and uncertainty of wildfire spread, instead of simply fitting the pixel average. Furthermore, the autoregressive framework facilitates systematic temporal forecasting of wildfire evolution. The proposed CGAN-based autoregressive framework enhances both the accuracy and physical interpretability of wildfire spread prediction, offering a promising foundation for time-sensitive response and evacuation planning. The supporting source code and data are available for download at [URL will be written when the paper is accepted].

## 1. Introduction

Wildfires are one of the most destructive natural disasters, posing a significant threat to both ecosystems and human life (Chen et al., 2024; Gajendiran et al., 2024). For instance, in March 2022, South Korea experienced its worst wildfire disaster in recorded history when fires broke out in Uljin County and Samcheok areas of Gangwon Province. The wildfire burned for nine consecutive days, consuming over 20,523 hectares of forest area, equivalent to one-third of Seoul's total area, and destroying 332 residential buildings, forcing 587 residents to evacuate. The economic losses from this disaster exceeded 908.6 billion Korean won (approximately $680 million USD) in direct damages (Kim et al., 2024). Similarly, while the comprehensive impact of the January 2025 Los Angeles wildfires is still under consideration, a report by Los Angeles County Economic Development Corporation estimates that these fires have destroyed more than 16,000 structures and cost an estimated $28 billion to $54 billion in economic losses (Horton et al., 2025).

The risks associated with wildfires are expected to escalate further as climate change drives rising global temperatures and prolonged droughts, contributing to more frequent and intense fire events (Carvalho et al., 2025). While preventing wildfires remains the most effective risk mitigation strategy, it is important to acknowledge that some wildfires are ignited by natural causes, such as lightning strikes (Pérez-Invernón et al.,2023). Therefore, once a wildfire has ignited, minimizing its impact becomes critical. This requires an efficient allocation of resources in both time and space, which can be achieved through the accurate prediction of wildfire spread patterns (Shadrin et al., 2024).

Substantial research efforts have been devoted to predicting wildfire spread. Existing approaches generally fall into two main categories: (1) empirical models (Brown et al., 2021; De Jong et al., 2016; Dolling et al., 2005; Jolly et al., 2015) and (2) physical modeling approaches (Mandel et al., 2011; Morvan & Dupuy, 2004; Mueller et al., 2021; Yoo et al., 2024). Empirical approaches, such as the fire weather index (FWI), utilize statistical techniques and historical weather data including temperature, humidity, and wind speed to assess the risk of fire occurrence (Van Wagner, 1987). When combined with spatial information systems, these approaches have been further developed to generate wildfire risk heatmaps (Fiorucci et al., 2008). On the other hand, physical models aim to simulate the underlying mechanisms of wildfire propagation over time. One of the most widely used methods is FARSITE, which incorporates detailed inputs such as weather conditions, terrain, and fuel characteristics, to predict the diffusion process of wildfires at specific locations and times (Finney, 1998).

Empirical models have been considered computationally efficient and valuable for large scale fire hazard rating systems. However, their usage has generally been limited in predicting the speed and direction of diffusion required in real-time fire diffusion situations (Di Giuseppe et al., 2025). Physics based models, e.g., FARSITE, have addressed this limitation by incorporating detailed formulations, including fuel moisture models and fire acceleration theories, and by leveraging extensive environmental datasets. The inclusion of such details, however, entails substantial computational costs. Furthermore, the methodological complexity of capturing highly uncertain phenomena, such as terrain induced wind shifts and sporadic spotting, hampers the operational deployment of such models for rapid, real-time decision-making required in emergency wildfire management (Sullivan, 2009)

To address these limitations, deep learning techniques have been introduced for wildfire prediction (Bhowmik et al., 2023; Jiang et al., 2023; Khennou & Akhloufi, 2023; Shadrin et al., 2024). A key advantage of these methods lies in their ability to directly learn complex spatiotemporal patterns from massive datasets without relying on predefined physical equations. A representative example is the model developed by Jiang et al. (2023), which employs an autoencoder (AE) architecture. In this framework, the encoder learns a compact representation of the current fire state, while the decoder reconstructs it into a future fire spread map. According to Jiang et al. (2023), this approach successfully achieved high pixel-level accuracy while reducing the computational burden of the physical model. However, AE-based models face fundamental limitations. Most rely on pixel-level loss functions such as mean squared error (MSE), which causes the model to produce a single deterministic output that represents the average of all possible future fire morphologies. This averaging effect often produces overly smooth or blurry predictions, thereby failing to capture the complex and irregular nature of real wildfire fronts.

This study proposes a wildfire spread prediction model based on an autoregressive conditional generative adversarial network (CGAN) to address this research gap. The contribution of the proposed model can be characterized as follows. First, by formulating wildfire spread prediction as an autoregressive task, the model reduces problem complexity and improves training stability. Second, by leveraging the adversarial learning process, the model overcomes the limitations of pixel-level loss functions, enabling it to capture the complex spatiotemporal dynamics of wildfire spread and to generate clear, realistic scenarios rather than blurred averages. Third, by incorporating ensemble techniques, the model accounts for aleatoric uncertainty in the input features, thereby enabling probabilistic wildfire spread prediction and promoting stable image generation. The effectiveness of the proposed CGAN-based autoregressive model is demonstrated through a comparative analysis with an existing AE-based model. The evaluation includes both qualitative and quantitative assessments: qualitatively, by visually comparing generated spread maps with actual wildfire imagery, and quantitatively, by employing metrics such as MSE and structural similarity index measure (SSIM), and boundary-based mean absolute error (BMAE).

The remainder of this paper is organized as follows. Section 2 reviews the background of FARSITE, a representative physical model, and the deep learning architectures of AEs and CGANs. Section 3 describes the proposed methodology, including the data generation, architecture of the CGAN-based autoregressive model, and data preprocessing procedures. Section 4 presents experimental results along with a comparative analysis against existing approaches. Finally, Section 5 concludes the paper by summarizing the research results and suggesting future research directions.

## 2. Background Knowledge

### 2.1. FARSITE

FARSITE is a 2D wildfire spread and behavior simulator developed for the U.S. Forest Service (Finney, 1998). It operates by integrating multiple validated physical sub models for phenomena such as surface fire, crown fire, and precipitation into a comprehensive predictive framework (Rothermel, 1972). By processing a wide range of environmental input data, including terrain features, fuel properties, and meteorological conditions, FARSITE generates detailed results, such as fire boundaries, spread rates, and flame lengths. This physics-based foundation enables high-fidelity predictions and is widely recognized for its accuracy in simulating fire behavior in complex terrain (Jahdi et al., 2015; Radočaj et al., 2022; Yoo & Song, 2025). However, the accuracy of FARSITE depends heavily on the availability of extensive and precise input data. Furthermore, computational complexity increases significantly with the scale and duration of the predictions, limiting the practicality of FARSITE for rapid decision-making in large-scale wildfire emergencies.

### 2.2. Autoencoder-Based Image Generation Model

An AE is a neural network architecture primarily used in unsupervised learning to derive efficient representations of data (Bank et al., 2023; Kim & Kim, 2022). It consists of two main components as shown in Figure 1: an encoder, which maps the input data to a low-dimensional latent representation, and a decoder, which reconstructs the original data from the latent space. AEs extract and reconstruct spatial features via an encoder–decoder pair, widely applied to image-to-image translation such as U-Net. A prominent example is the U-Net architecture, commonly used for image generation (Ronneberger et al., 2015). Training of such models is typically guided by a pixel-wise loss function, such as MSE, which quantifies the difference between the generated output and the target image.

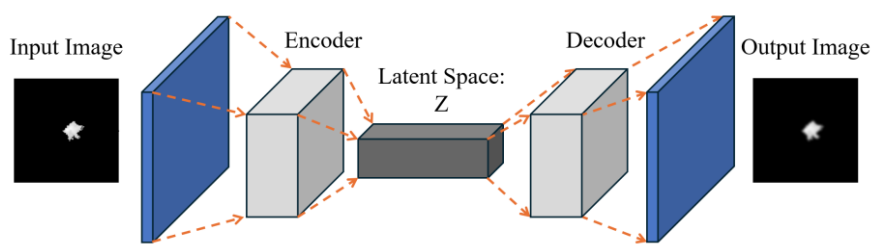

**Figure 1.** A schematic diagram of an autoencoder.

### 2.3. Conditional Generative Adversarial Network

A generative adversarial network (GAN) consists of two neural networks: a generator and a discriminator. As shown in Figure 2, the generator receives random vectors from the latent space as input and generates data that closely resembles the distribution of real data. The discriminator, on the other hand, acts as a binary classifier, accepting both real data and data generated by the generator as input and determining whether the data is real or generated. Both networks are trained in a competitive minimax game theory (Goodfellow et al., 2014). Through this adversarial competition, the discriminator becomes increasingly adept at distinguishing between real and synthetic data, leading the generator to produce increasingly high-quality outputs capable of fooling the discriminator.

CGANs are an extension of the original GAN framework by introducing conditional inputs to both the generator and discriminator to control the generation process (Mirza & Osindero, 2014). While standard GANs generate output solely based on random noise, CGANs incorporate conditional information such as class labels, text, or, as in this study, environmental variables like terrain and weather to guide the generation process. The generator learns to produce data consistent with these conditions, while the discriminator evaluates whether the generated data aligns with the specified conditions.

A key advantage of CGANs lies in their flexibility for modeling complex and nonlinear data sets. Compared to pixel-wise losses, the discriminator in a CGAN serves as a learnable, data-driven loss function. This dynamic process prevents the generator from producing the average and blurred outputs that characterize the limitations of AE-based models. By encouraging the generator to produce realistic outputs that are statistically indistinguishable from real data, CGANs offer methodological advantages particularly suited to predicting specific wildfire spread patterns from complex environmental inputs (Isola et al., 2017).

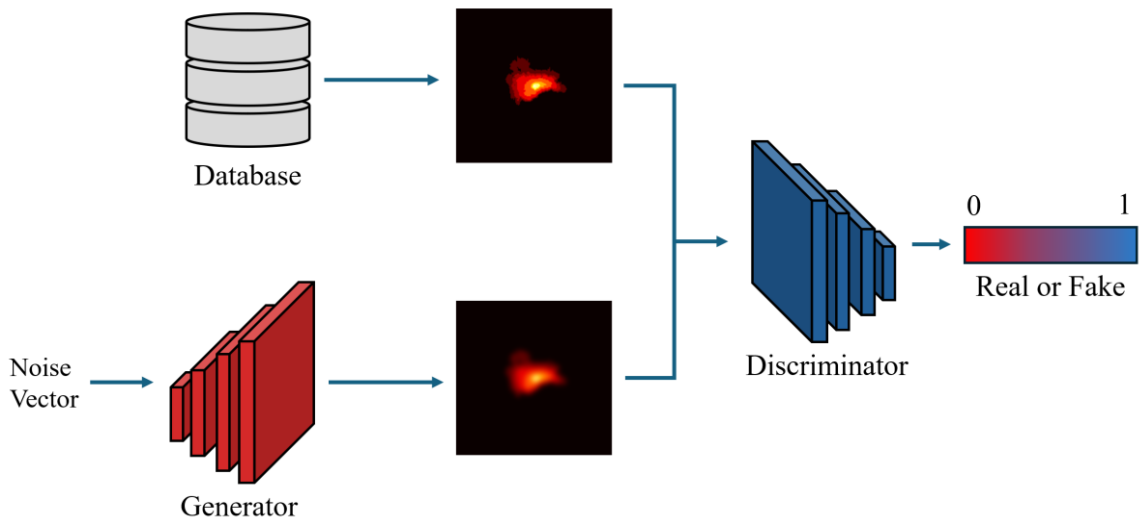

**Figure 2.** A schematic diagram of generative adversarial network.

## 3. CGAN-Based Wildfire Spread Prediction Model

This section provides a detailed account of the methodology proposed for predicting wildfire spread. Section 3.1 introduces the autoregressive prediction technique, which reduces the complexity of modeling wildfire dynamics. Section 3.2 presents the architecture of the proposed prediction model and outlines its key components. Section 3.3 describes the construction of the dataset used for training, while Section 3.4 details the model's training procedure and prediction strategy.

### 3.1. Problem Formulation: Autoregressive Prediction

Most existing studies have focused on predicting wildfire spread patterns (or the final fire boundaries) at a specific time point (Jiang et al., 2023; Wu et al., 2022; Yoo et al., 2024). However, for practical decision-making related to evacuation planning and resource allocation (e.g., firefighting and water deployment), it is equally important to predict not only the final fire extent but also the intermediate spread patterns over time. To address this need, this study formulates wildfire prediction as an autoregressive problem, in which the task is to learn a transition function $f$ that predicts the future wildfire state $S_{t+\Delta t}$ based on the current state $S_t$ and the corresponding environmental conditions $C_t$ (Thedy et al., 2025):

$$S_{t+\Delta t} = f(S_t, C_t) \tag{1}$$

where $\Delta t$ is the prediction time interval. By iteratively applying this transition function, a sequence of future wildfire spread states $\{S_1, S_2, \dots, S_n\}$ can be generated based on the initial state $S_0$ and the associated time-varying environmental conditions $\{C_0, C_1, \dots, C_{n-1}\}$ (e.g., meteorological data). A schematic overview of this autoregressive prediction mechanism applied to wildfire prediction is presented in Figure 3.

This autoregressive formulation offers several key advantages over a one-shot approach that directly maps initial conditions to a final wildfire spread pattern. First, it enables the model to learn the fundamental dynamics of state transitions, instead of a single, highly complex correlation between an input and an eventual outcome. This stepwise learning improves prediction stability and reduces the risk of catastrophic error accumulation in long-range for[ecasts. Second, the autoregressive framework naturally produces the full sequence of intermediate fire states. This sequential information is of substantial practical value, as it provides emergency responders with time-resolved insights into the evolution of wildfire spread, thereby supporting more effective and timely decision-making in disaster management.

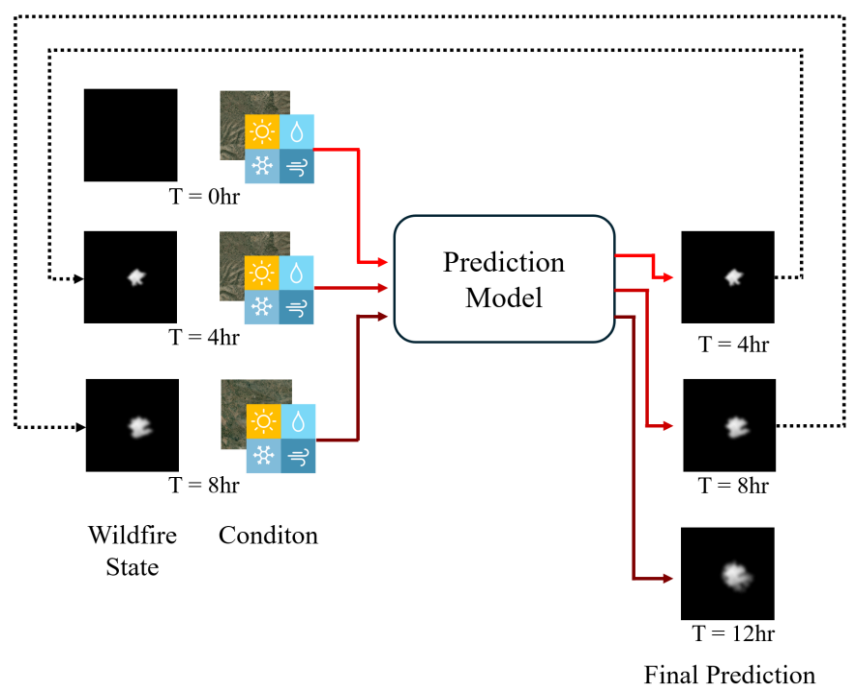

**Figure 3.** Autoregressive prediction mechanism.

## 3.2. Proposed CGAN-Based Autoregressive Model

### 3.2.1. Overall Architecture

Figure 4 illustrates the proposed CGAN architecture. The generator performs the autoregressive prediction step by taking the current wildfire spread state ($S_t$), environmental condition ($C_t$), and a random noise vector ($z$) as inputs. It then produces a prediction of the subsequent step ($\hat{S}_{t+1}$). Note that $z$ is randomly sampled from the standard normal distribution. The discriminator assesses the plausibility of the prediction image ($\hat{S}_{t+1}$) comparing it against true wildfire data ($S_{t+1}$), conditioned on the environmental variables at the corresponding time step ($C_{t+1}$). Through this adversarial process, the generator is guided to produce increasingly realistic wildfire spread patterns.

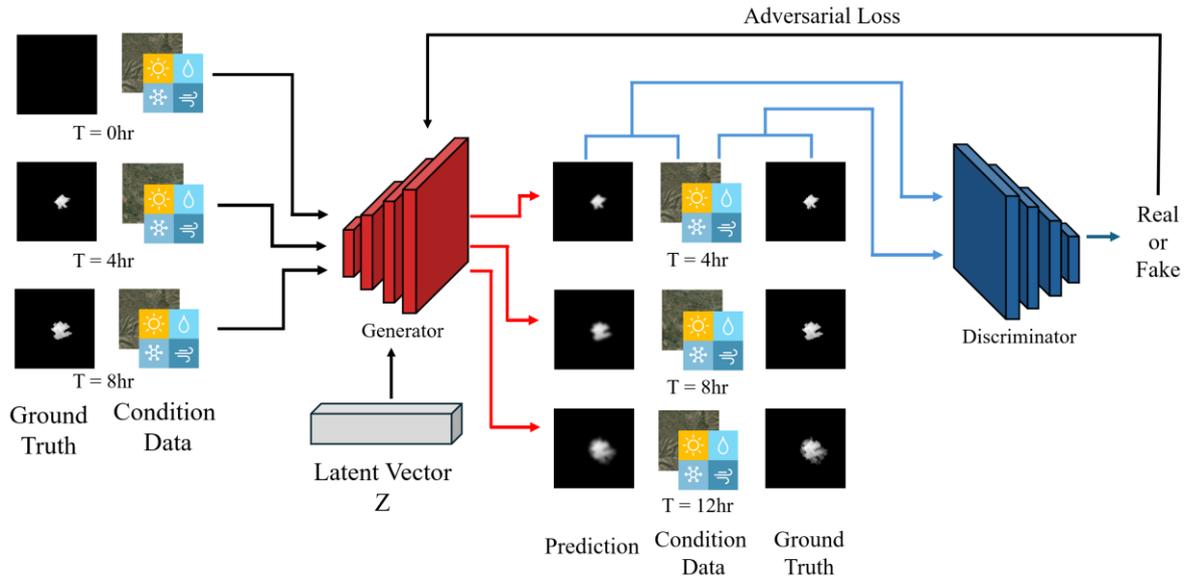

**Figure 4**. Proposed CGAN-based autoregressive model for wildfire prediction.

A key feature of the proposed model is the incorporation of the random noise vector ($z$) as an input, which allows the model to generate different outcomes for each prediction. The introduction of the noise vector provides two main advantages: (1) it stabilizes the training of the autoregressive CGAN-based model, and (2) it enables probabilistic prediction. First, regarding training stability, we observe that without the noise vector the model fails to generate realistic wildfire spread patterns. Specifically, it either produces poorly reconstructed outputs or exhibits artifacts such as checkerboard patterns, even when the loss function appears to have converged. This indicates that the absence of stochastic variability restricts the generator's learning capacity, leading to degenerate solutions. The noise vector mitigates this issue by injecting controlled randomness into the generation process, which enhances diversity in the input and prevents collapse into trivial solutions. Second, the stochastic property of the noise vector enables probabilistic forecasting of wildfire spread at the grid (or pixel) level. For instance, under a given set of environmental conditions, the model can generate a large ensemble of predictions (e.g., 100 simulated wildfire spread patterns). By counting the number of times each grid cell is classified as burned across the ensemble, it becomes possible to estimate the probability of fire occurrence in that location. This framework offers a principled means of accounting for aleatoric uncertainty, i.e., the inherent variability in wildfire spread that cannot be fully captured by input data. Such ensemble-based probabilistic predictions have also been well-established in existing literature (Dacre et al., 2018; Kim et al., 2020; Wang et al., 2024). Importantly, while details on runtime performance are discussed in Section 4, the computational cost per prediction step is on the order of one second with Intel i9-14900K, 64GB system memory and RTX 3060 12GB. Thus, generating multiple predictions with varying noise vectors for probabilistic forecasting does not impose prohibitive computational demands.

### 3.2.2. Generator

The generator is designed to execute the single-step autoregressive prediction, mapping the current state ($S_t, C_t$) to the subsequent state $\hat{S}_{t+1}$. The detailed architecture of the generator is presented in Figure 5. The process begins with extracting features from the current wildfire spread state, where spatial representations are learned through a dedicated fire feature extractor. In parallel, conditional inputs, including static terrain data and dynamic meteorological variables (i.e., wind, temperature, and relative humidity), are incorporated. The terrain data, which contain spatial information, are processed by a separate terrain feature extractor. This feature extractor is implemented as a convolutional neural network consisting of two hidden layers without down sampling and a batch normalization layer. Note that we refer to the information processed through each feature extractor as a feature map. Also note that a description of the dimensional properties of each input data type, as represented in Figure 5, will be provided in Section 3.3.

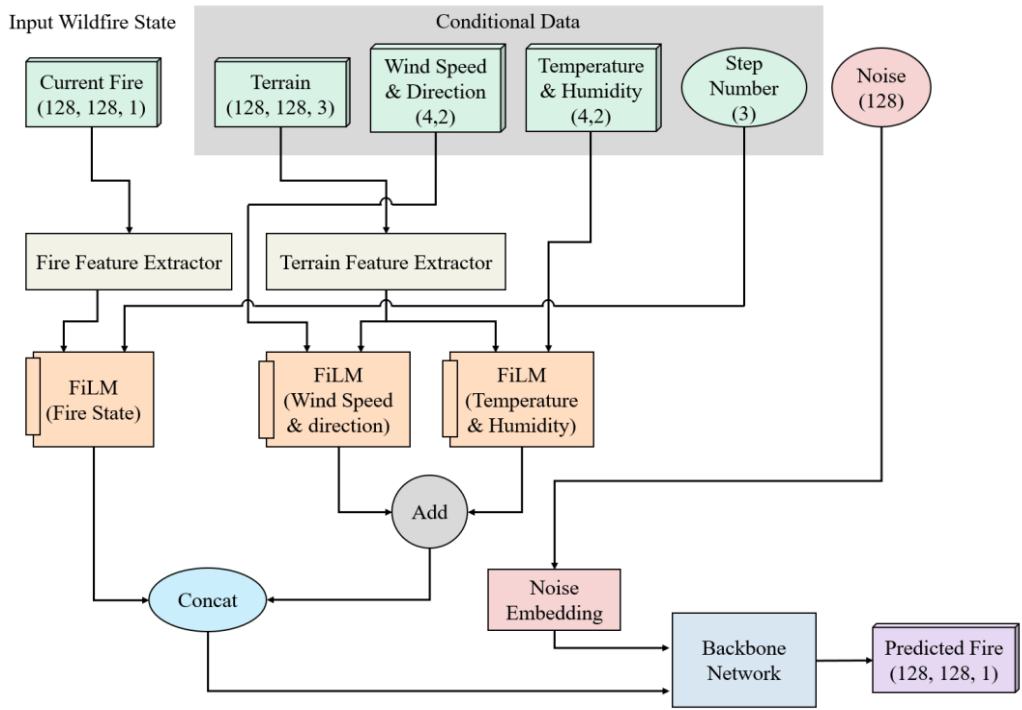

**Figure 5.** The overall architecture of the proposed generator, where environmental inputs (terrain, wind, temp/humidity) are processed and injected into the fire feature map using FiLM layers, which is then concatenated with a latent noise vector and fed into the backbone U-Net. "Concat" refers to the concatenation operation.

To effectively fuse the non-spatial environmental conditions with the preprocessed spatial feature information (i.e., feature maps), we employ feature-wise linear modulation (FiLM) layers (Perez et al., 2018). The primary advantage of FiLM layers is their ability to prevent overly large model sizes or the dominance of one input modality when merging heterogeneous data sources. The operational mechanism of a FiLM layer is illustrated in Figure 6(a). For a given feature map, the FiLM layer predicts a channel-wise scaling parameter $\gamma$ and a shifting parameter $\beta$ from the condition vector using two separate fully connected networks. These parameters are then applied via an affine transformation of the feature map, thereby allowing the environmental conditions to dynamically modulate feature extraction and guide the generation process across multiple scales.

After preprocessing each input, the generated feature maps are passed to the backbone U-Net along with the latent noise vector ($z$), which is responsible for synthesizing the predicted wildfire spread. The detailed architecture of the U-Net is shown in Figure 6(b). The encoder path consists of a series of residual blocks that progressively down-sample the input, thereby learning hierarchical feature representations at multiple spatial scales. Symmetrically, the decoder path up-samples these abstract representations while incorporating skip connections (He et al., 2016). These skip connections concatenate encoder feature maps directly with their corresponding decoder layers, enabling the network to effectively recover fine-grained spatial details that are essential for high-resolution predictions. A key architectural element is the injection of the latent noise vector ($z$), which provides the stochasticity required for probabilistic prediction, as discussed in Section 3.2.1. This is implemented at the bottleneck of the U-Net, where the noise vector is reshaped into a tensor and concatenated with the feature map.

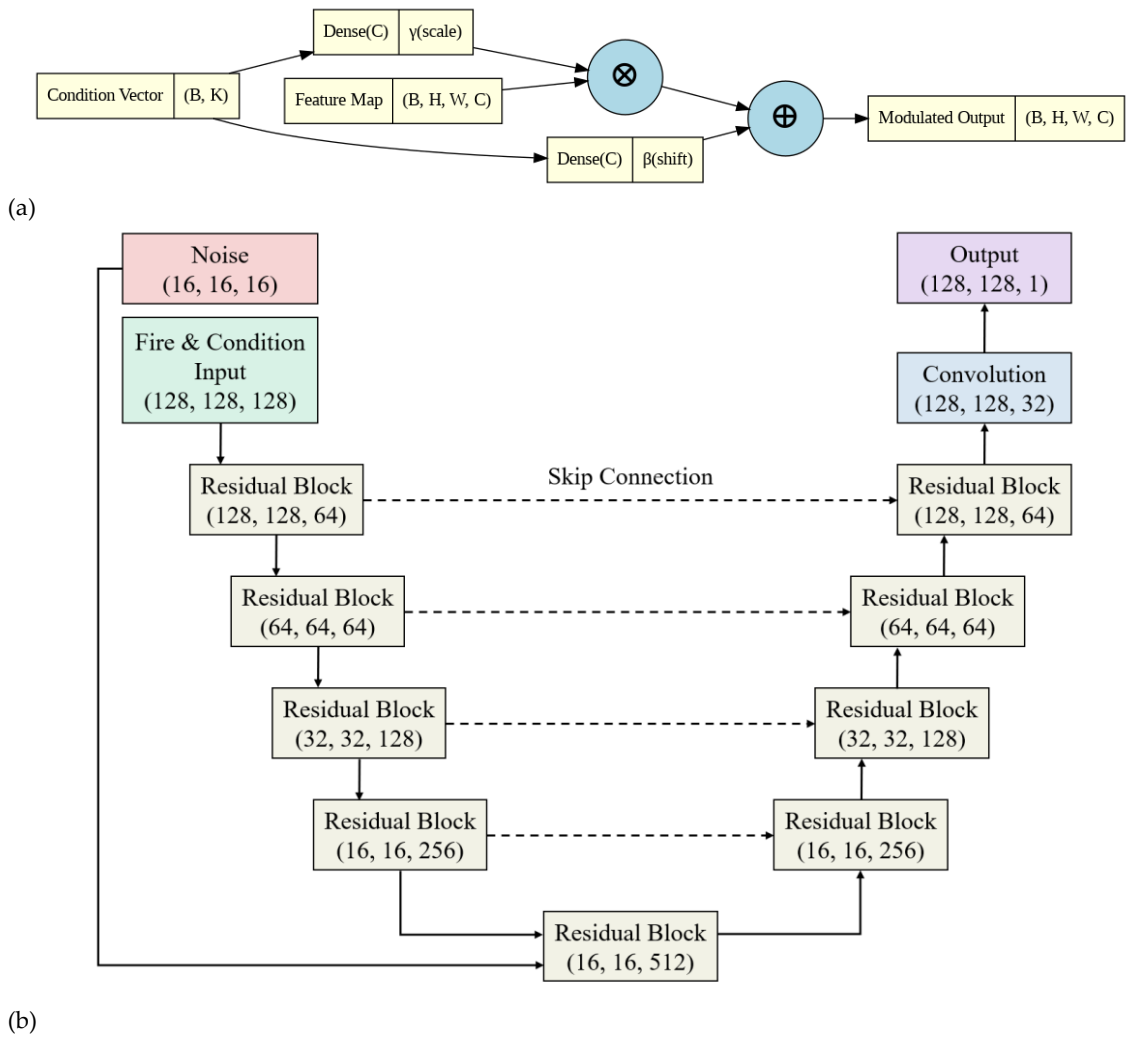

(a)

(b)

**Figure 6.** Detailed description of the techniques used: (a) FiLM layer structure for feature embedding, and (b) the detailed architecture of the backbone U-Net, showing the downsampling path with residual blocks, the injection of the noise tensor at the bottleneck, and the upsampling path with skip connections to generate the final fire map.

### 3.2.3. Discriminator

The discriminator serves as an adversary to the generator during training, with the primary objective of distinguishing between real data pairs $(S_{t+1}, C_t)$, and predicted data pairs $(\hat{S}_{t+1}, C_t)$. It provides feedback, in the form of adversarial loss, that guides the generator toward producing more realistic outputs. The detailed structure of the discriminator is presented in Figure 7. Similar to the generator, the model first processes and fuses various input information, including fire maps and conditional data (terrain, weather, and time intervals), using feature extractors followed by FiLM layers. A key distinction, however, is that the discriminator does not incorporate a noise vector, since its role is to reliably determine the authenticity of the generated wildfire spread patterns rather than introduce stochasticity.

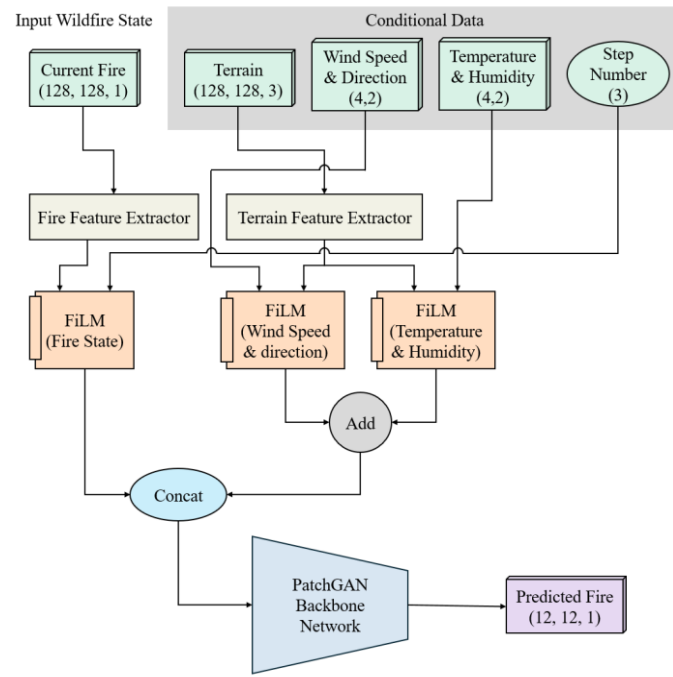

**Figure 7.** The architecture of the discriminator. "Concat" refers to the concatenation operation.

After preprocessing, the fused feature representation is fed into the PatchGAN style backbone network (Isola et al., 2017), which consists of a series of stride convolutional layers. Compared to a conventional discriminator that outputs a single scalar value to indicate the authenticity of an entire image, PatchGAN evaluates the authenticity of regions of local image by partitioning the input into overlapping patches, as shown in Figure 8. In particular, the discriminator applies a fully convolutional computation to the input image, producing an $N \times N$ output grid, where each element corresponds to a specific patch and represents the probability that the patch is real. In this study, $N$ is set to 12. The final discriminator loss (i.e., $\mathcal{L}(D)$ in Figure 8) is calculated by averaging the classification results across all patches.

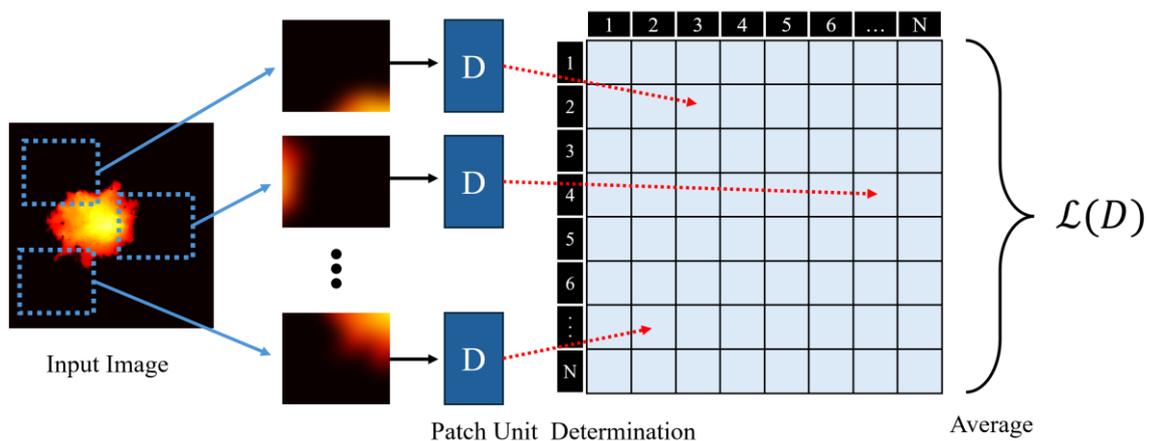

**Figure 8.** PatchGAN's image determination process. "D" refers to a discriminator.

### 3.2.4. Objective Function

The training of the proposed model is guided by a dual-objective optimization process, with distinct objective functions for the generator ($G$) and the discriminator ($D$). The following structure ensures stable and effective adversarial training.

To formalize the objective functions, we first define our notation. Let $x$ represent the corresponding ground truth fire map, $y$ be the set of conditional inputs (i.e., current wildfire spread state, terrain image, and meteorological data), and $z$ be a random noise vector. The generator produces a predicted wildfire spread state $\hat{x} = G(z, y)$, and the discriminator outputs a real-valued score for a given pair of inputs and a wildfire spread state. The operator $E[\cdot]$ denotes the expectation, which is the average value computed over a batch of data samples.

The generator's performance is optimized using a composite objective function, defined as a weighted sum of three distinct components. This multi-faceted loss function is designed to balance structural accuracy with perceptual quality in the generated predictions. The three components are as follows.

- Adversarial loss: This guides the generator to produce outputs indistinguishable from real data by the discriminator. The goal is to maximize the discriminator's score for the generated data, which is achieved by minimizing the negative values of that score.

$$\mathcal{L}_{adv} = -E_{\hat{X},Y}[D(\hat{x}, y)] \tag{2}$$

- $L_1$ loss: This loss measures the pixel-wise mean absolute error between the generated and true images. It enforces structural fidelity by ensuring that the generated fire spread map aligns with the ground truth in terms of shape and location. Prior work has shown that combining adversarial and $L_1$ loss leads to sharper and more accurate image predictions (Isola et al., 2017).

$$\mathcal{L}_{L1} = E_{X,\hat{X}}[\parallel x - \hat{x} \parallel_1] \tag{3}$$

- Dice loss: Widely used in image segmentation, the dice loss maximizes the overlap between the predicted and actual fire areas. It is particularly effective when the burned area constitutes a small fraction of the overall map, as it directly emphasizes region-level similarity (Milletari et al., 2016). Here, $\varepsilon$ represents a very small value, such as 1e-7, to prevent division by zero.

$$\mathcal{L}_{Dice} = 1 - \frac{2\sum_{i=1}^{N} x_i \hat{x}_i + \varepsilon}{\sum_{i=1}^{N} x_i + \sum_{i=1}^{N} \hat{x}_i + \varepsilon} \tag{4}$$

The overall generator loss is a weighted sum of these three components:

$$\mathcal{L}_G = w_{adv}\mathcal{L}_{adv} + w_{L1}\mathcal{L}_{L1} + w_{Dice}\mathcal{L}_{Dice} \tag{5}$$

The discriminator is optimized using a loss function adopted from Gulrajani et al. (2017). This function is designed to provide a more stable training process compared to the original GAN. It consists of two parts:

- Wasserstein loss: This term encourages the discriminator to output higher scores for real data than for generated data. $\mathcal{L}_{WGAN} = E_{\hat{X},Y}[D(\hat{x}, y)] - E_{X,Y}[D(x, y)]$

$$\mathcal{L}_{WGAN} = E_{\hat{X},Y}[D(\hat{x}, y)] - E_{X,Y}[D(x, y)] \tag{6}$$

- Gradient penalty: It imposes a penalty on the discriminator's gradient norm to enforce the 1-Lipschitz constraint, ensuring training stability. The penalty is applied to interpolated samples ($\hat{x}$) that lie between real and generated images.

$$\mathcal{L}_{GP} = E_{\hat{X},Y}[(||\nabla_{\hat{x}}D(\hat{x}, y)||_2 - 1)^2] \tag{7}$$

The final discriminator loss is a combination of these two components:

$$\mathcal{L}_D = \mathcal{L}_{WGAN} + w_{GP}\mathcal{L}_{GP} \tag{8}$$

where $w_{adv}$, $w_{L1}$, $w_{Dice}$, and $w_{GP}$ are weights that control the relative importance of adversarial loss, $L_1$ loss, dice loss, and gradient penalty loss, respectively. These weights are empirically determined through iterative training to achieve an optimal balance between prediction accuracy and segmentation quality. In this study, $w_{adv} = 1.0, w_{L1} = 20, w_{Dice} = 0.3$, and $w_{GP} = 10$.

3.3. Construction of Dataset

To train and evaluate the proposed CGAN-based prediction model, we construct a large-scale hypothetical wildfire dataset using FARSITE. Northern California, USA, is selected as the study area, given its high wildfire frequency and relevance to prior research in wildfire modeling (Airey-Lauvaux et al., 2022; Yoo & Song, 2025; Zhou et al., 2020). Moreover, previous studies have demonstrated that FARSITE provides accurate predictions of wildfire spread in this region when compared against observed fire perimeters (Fraga et al., 2024; González et al.,2025; Westerling, 2016). While the proposed model can be trained with various prediction horizons, in this study, we focus on three sequential steps spanning a total of 12 hours (i.e., 4-, 8-, and 12-hour). Accordingly, the dataset is constructed to match this temporal structure. While this study focuses on building an effective prediction model for a 12-hour period, future research should consider extending the prediction period.

The dataset consists of three main components: (1) meteorological data, (2) topographic data, and (3) wildfire spread imagery generated by FARSITE. Meteorological and topographic data serve as conditional inputs to the CGAN-based model, while the wildfire spread imagery provides the ground-truth target for training. To incorporate realistic fire-prone conditions, data are collected for the wildfire season (July to September) from 2016 to 2020 based on statistical reports identifying this as the peak wildfire period in California (CAL FIRE, 2021). Meteorological data are obtained via the Open-Meteo historical weather API, while topographic data are derived from Sentinel-2 satellite imagery processed through QGIS (Zhu & Woodcock, 2014).

The study area is defined as the Northern California badlands. To generate a diverse range of fire spread cases, we first designate a 6-kilometer square area from which an ignition point is selected, with the size determined by the effective range of the meteorological data. Within this zone, 300 initial ignition points are randomly selected. For each ignition point, a dedicated 8-kilometer square simulation site is defined, centered on the respective point. The selection of this dimension is linked to the spatial resolution of our source meteorological data, the ECMWF IFS dataset, which has a native grid resolution of 9 km. By setting our simulation domain slightly smaller than a single weather grid cell, we ensure that the atmospheric conditions driving each simulation are spatially coherent and representative.

For each site, 70 distinct weather scenarios of 12-hour duration are employed based on the collected meteorological data. This combination results in an initial dataset of 21,000 wildfire simulations (300 fire sites × 70 weather conditions). Simulations where the fire spread extended beyond the study boundaries are excluded, yielding a final dataset of 12,604 valid samples for model training. This filtering ensured that only meaningful wildfire spread events are retained, enhancing both learning efficiency and practical applicability. However, due to this process, the model trained with this dataset has a limitation, in that it cannot predict large-scale wildfires that exceed the target boundary.

To incorporate auto-regressive characteristics, the FARSITE simulation outputs (ground truth values ranging from 0 to 720 minutes that represent the fire arrival time at each cell) are converted into single-channel grayscale images. Pixel intensities are linearly scaled from 255 (earliest ignition, 0 minute) to 30 (latest ignition, 720 minutes), and partitioned into three equal intervals corresponding to 4-, 8-, and 12-hour wildfire spread maps. The simulation outputs from FARSITE are resized to 128 × 128 pixels, corresponding to a spatial resolution of 62.5 m per pixel. Each wildfire spread image is paired with a corresponding three-channel (RGB) terrain image derived from satellite data. For the meteorological inputs, four variables such as wind speed, wind direction, temperature, and relative humidity, are aligned with the three predictive time steps. Each step incorporates four hours of data for these variables, resulting in a structured meteorological input tensor with dimensions (3, 4, 4). A schematic overview of the dataset generation procedure is provided in Figure 9.

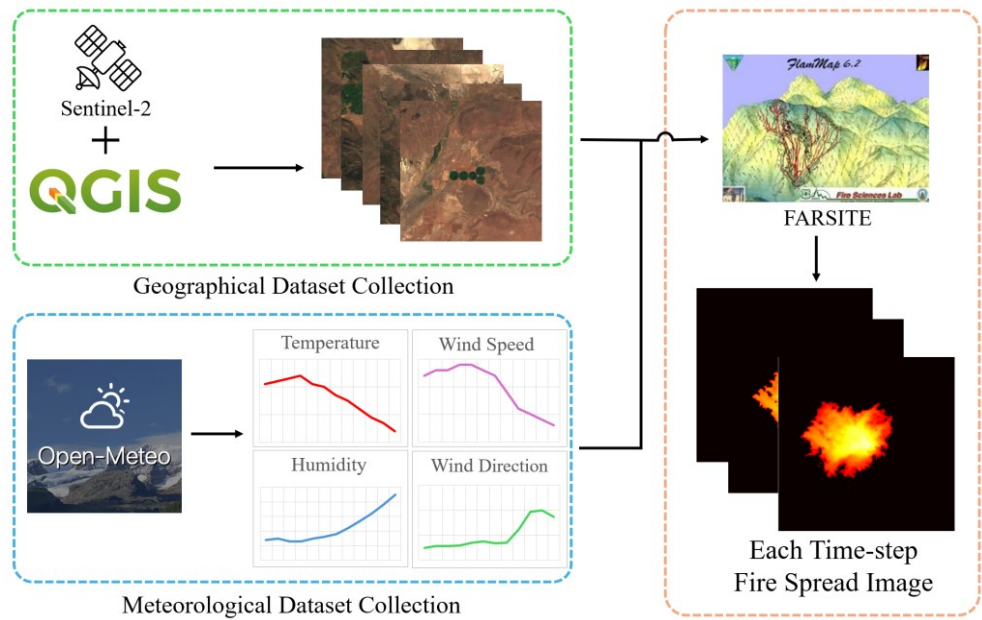

**Figure 9**. Dataset generation framework.

### 3.4. Training and Prediction Strategies

Using the constructed dataset, the model is trained on randomly selected pairs of consecutive states $(S_t, S_{t+\Delta t})$, along with the corresponding conditional data $(C_t)$, rather than strictly consecutive triplets of the three predefined time steps. This strategy is adopted because, during training, the generator does not rely on its own previously generated predictions but instead uses the true wildfire spread images as input. By shuffling time steps and sampling pairs randomly, the model is exposed to a wider range of wildfire states. This maximizes sample diversity and enables the CGAN-based model to more effectively learn the fundamental dynamics of wildfire evolution. During prediction, however, the trained model uses its previously predicted spread image as input to predict the subsequent time step, consistent with the autoregressive framework.

For model training, 80% of the dataset (10,083 samples) is used for training, while the remaining 20% (2,521 samples) serves as a test set for validation. Training is conducted on a high-performance workstation equipped with 384 GB of RAM and two NVIDIA RTX A5000 GPUs. With a batch size of 16, the training process requires approximately 21 hours to complete 1,500 epochs, at which no further improvement in model performance is observed. The Adam optimizer is employed to update model parameters.

During prediction, the model operates within the autoregressive loop described in Section 3.1 to predict a full sequence of future wildfire states. The process begins with an initial state $S_0$, and the predicted output at each step is recursively used as the input for the subsequent prediction. To enhance stability and robustness, an ensemble sampling strategy is employed at each prediction step, as illustrated in Figure 10. Given an input $(S_t, C_t)$, the generator produces $N_E$ different prediction samples $\{\hat{S}_{t+1}^1, \dots, \hat{S}_{t+1}^{N_E}\}$, by drawing from distinct latent noise vectors $\{z_1, \dots, z_{N_E}\}$. These predictions are, then, averaged to produce a single robust estimate $\hat{S}_{t+1}$, which represents the most probable fire spread pattern for the next time step. Note that $N_E$ is set as 5, in this study. This averaged map subsequently serves as the input for the next iteration of the loop. The ensemble prediction and averaging procedure continues until the final forecast horizon is reached. Depending on the decision-making context, however, the ensemble results need not rely solely on the mean value. Decision-makers with risk-averse or risk-neutral preferences may instead consider percentile-based estimates (e.g., the 10th or 90th percentile) to represent conservative or aggressive spread scenarios. The full source code for the proposed model and the test dataset subset are available at [URL will be written when the paper is accepted].

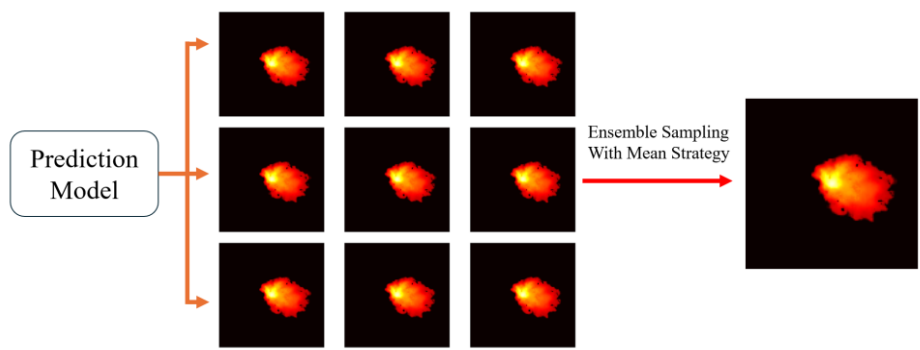

**Figure 10**. Ensemble sampling process for probabilistic prediction.

## 4. Numerical Investigation

### 4.1. Comparative Model: AE-Based Model

For comparative analysis, an AE architecture adapted from Jiang et al. (2023) is used as a baseline model. The structure of this model is illustrated in Figure 11. The core idea is to approximate the complex temporal evolution of wildfire spread as a single image-to-image transformation task, directly mapping environmental conditions to a predicted fire extent at a specified future time point. To achieve this, the model integrates several critical physical variables: (i) topographic data represented by a digital elevation model (DEM), capturing elevation and slope, (ii) fuel model data describing the type and density of combustible materials, and (iii) time-varying meteorological variables such as wind, temperature, and humidity. These heterogeneous inputs are fused through an internal module to form a unified feature representation, which is subsequently processed by a deep encoder–decoder network.

A critical adjustment is introduced to ensure fairness in the comparative analysis. The original model described by Jiang et al. (2023) incorporated an intermediate fire spread image (e.g., a 3-hour spread map when predicting a 12-hour outcome), referred to as a "fire-state," as an additional input. While this design can improve predictive accuracy, it risks artificially inflating performance by providing partial knowledge of the ground truth. More importantly, it does not reflect the operational reality of rapid initial response, where no intermediate fire spread map is available immediately following an outbreak. To align with the objectives of this study, the AE-based model is modified to exclude the fire-state input. Predictions are instead generated solely from environmental data and ignition point information.

The original AE-based model by Jiang et al. (2023) was designed to predict the final stage of wildfire spread. To compare the results, this study creates prediction images for each time period using pixel values corresponding to 4-, 8-, and 12-hour, based on a preprocessing strategy used when creating the dataset. The AE-based model is trained using a conventional approach that minimizes the MSE between the predicted and actual spread maps, as detailed in Jiang et al. (2023). The model is trained using the same training environment and dataset used to train the CGAN-based model. The full source code and detailed configuration settings for this model are available at [URL will be written when the paper is accepted].

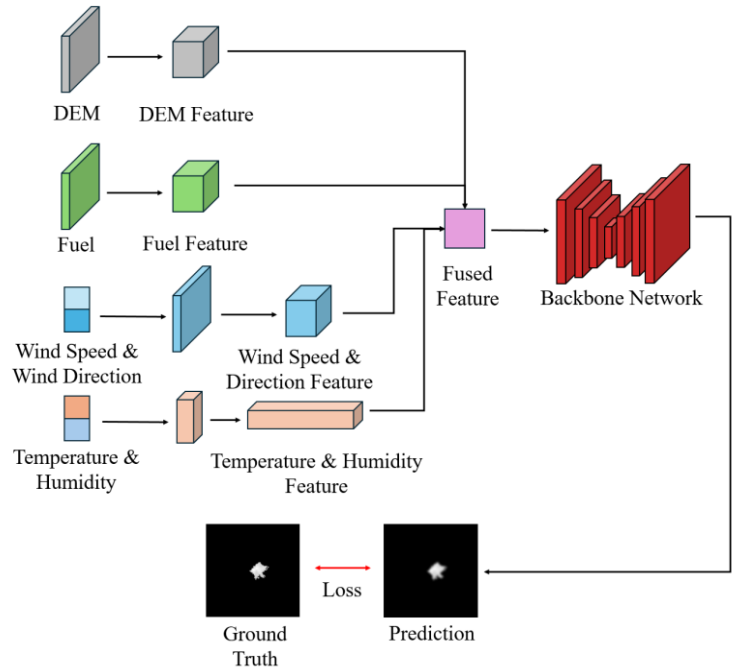

**Figure 11**. AE-based model overview.

## 4.2. Prediction Performance

Figure 12 compares the performance of the proposed CGAN-based model and the AE-based baseline in terms of MSE and SSIM, which evaluate overall prediction accuracy and the preservation of meaningful spatial patterns in wildfire spread, respectively. The MSE is defined as the average squared difference between predicted and ground-truth pixel values, with lower values indicating closer alignment between the prediction and the reference image. By contrast, SSIM quantifies structural similarity by comparing luminance, contrast, and spatial structure between two images. SSIM scores range from 0 to 1, with higher values reflecting stronger preservation of spatial integrity and image structure. Both metrics are calculated solely within a synthetic mask that combines the burned areas of the actual and predicted images. This approach focuses on areas where wildfires have occurred or are predicted to occur, excluding background pixels without fire information.

Across all timescales, the CGAN-based model consistently exhibits lower MSE values than the AE-based model. The CGAN-based model maintains temporal consistency and reconstructs fire patterns with higher spatial fidelity. This suggests that it more effectively captures the nonlinear dynamics of wildfire spread than the AE-based model trained solely on pixel-wise loss. However, the SSIM scores are similar between the two models. The reason for this skew is discussed in detail in Section 4.3. Jiang et al. (2023) reported significantly higher accuracy with an AE-based framework, but this is primarily due to the incorporation of intermediate fire spread images as input. However, as discussed earlier, this artificially inflates prediction accuracy and fails to reflect the operational constraints of rapid initial response scenarios, where such intermediate data is unavailable.

To more comprehensively evaluate each model's ability to delineate the fire boundary, Figure 13 presents the BMAE results. Boundary masks are extracted using boundary detection from both the actual and predicted fire areas, and the comparison is performed in terms of the mean absolute error (MAE). This choice is made because the squared error measure can exaggerate the impact of small pixel outliers along the boundary, whereas the MAE provides a more reliable and interpretable measure of local boundary deviation.

The AE-based model exhibits relatively large boundary errors over time, while the CGAN-based model maintains significantly smaller values at all prediction stages. This difference is greatest around 8-hour, indicating that the CGAN-based model better represents the rapid and irregular expansion of the fire front during the intermediate spread phase. Even at 12-hour prediction, the CGAN-based model continues to perform well, generating continuous and realistic fire boundaries that closely match the actual fire area. These results demonstrate that the adversarial learning framework enhances the model's ability to generate clear and physically plausible fire boundaries, rather than overly smoothed averages.

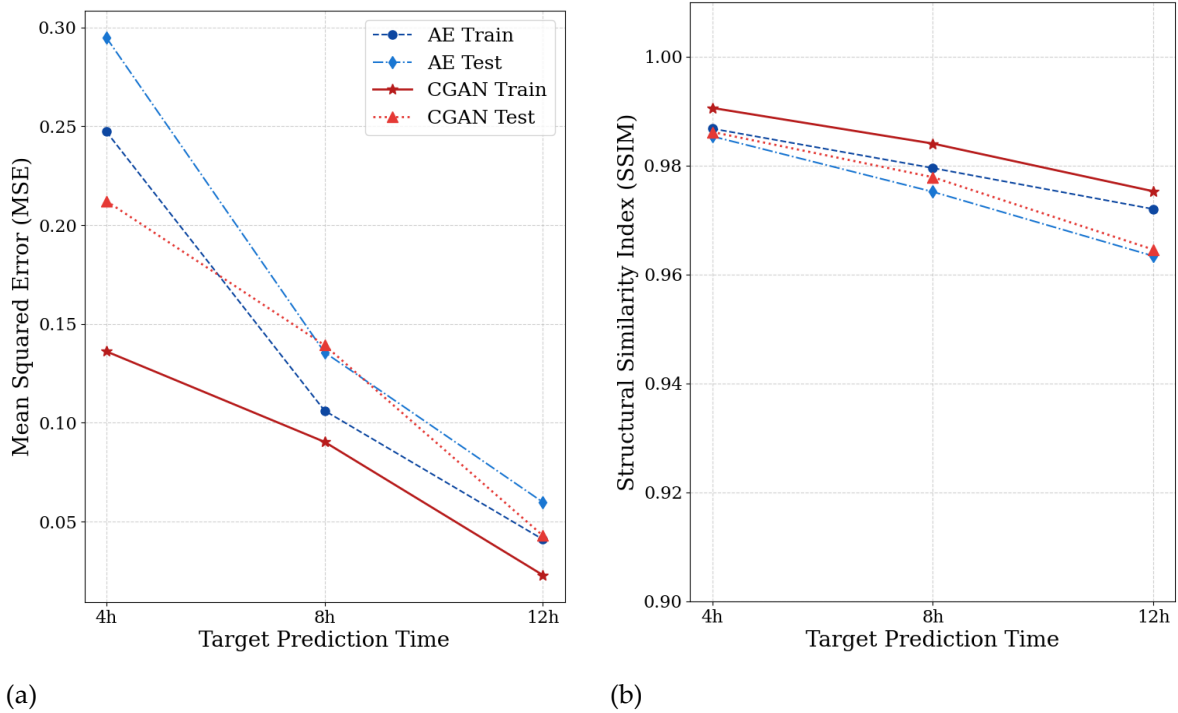

(a)

(b)

**Figure 12**. Performance measure graph (MSE, SSIM) between the two models: (a) comparison in terms of MSE, (b) comparison in terms of SSIM score.

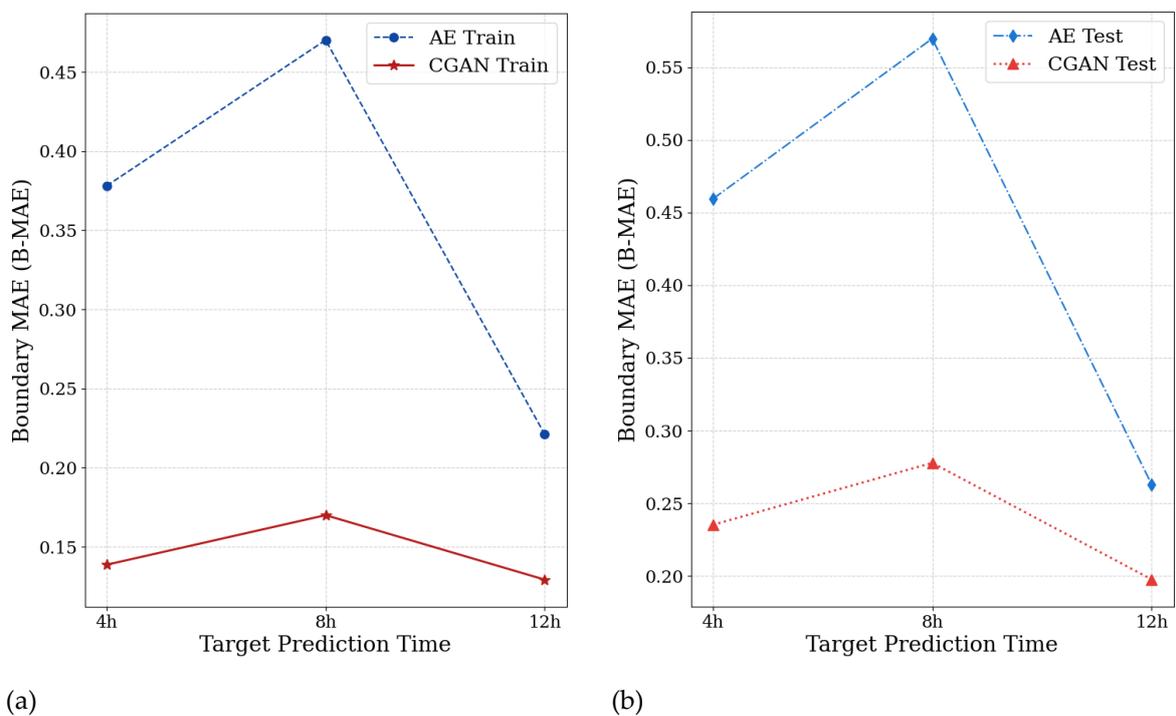

(a)

(b)

**Figure 13**. Performance measure graph (BMAE) between the two models: (a) comparison about the train dataset, (b) comparison about the test dataset.

Figure 14 presents a visual comparison of prediction results for four representative test samples. The first row displays the ground-truth wildfire spread at 4-, 8-, and 12-hour, while the second and third rows show the corresponding predictions generated by the CGAN-based model and the AE-based model, respectively. An examination of the AE-based predictions (third row) shows that the model can capture the general direction and approximate location of the wildfire spread. However, a clear limitation emerges in its representation of the fire perimeter. In contrast to the irregular and fragmented boundary of the actual fire, the AE-based outputs become progressively smoother and more rounded, particularly at the 12-hour prediction. This smoothing effect is a well-known consequence of optimizing pixel-wise loss functions such as MSE, as it targets the average value across the image. As a result, the AE-based predictions appear blurry and fail to reproduce the fine-grained textures and structural complexity of wildfire boundaries. In contrast, the predictions from the CGAN-based model (second row) exhibit a qualitatively different characteristic. Not only does the model capture the overall spread pattern, but it also reconstructs the fire perimeter with substantially greater sharpness and complexity. The generated boundaries closely resemble the irregular and textured morphology of the ground-truth fire fronts, indicating an ability to preserve fine structural details that are critical for operational decision-making. This improvement arises directly from the adversarial learning framework, which forces the generator to produce outputs that are both statistically plausible and perceptually convincing, thereby addressing the averaging limitations of purely pixel-wise optimization.

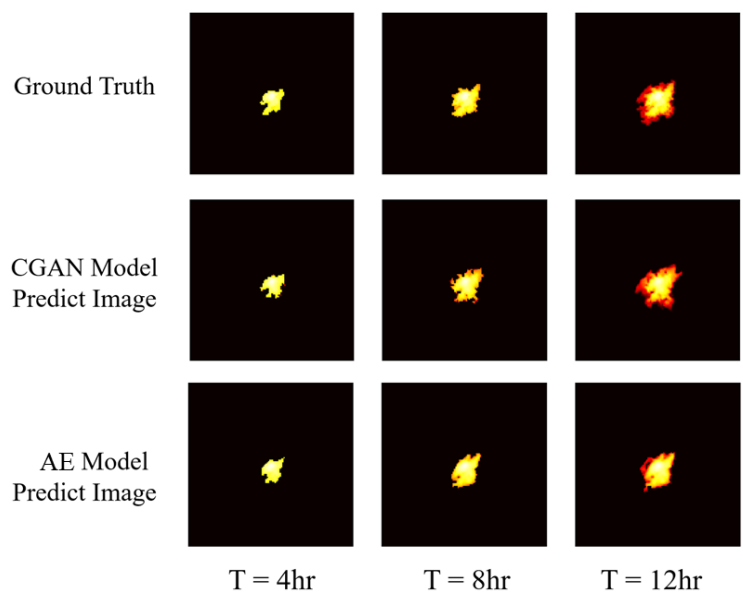

(a)

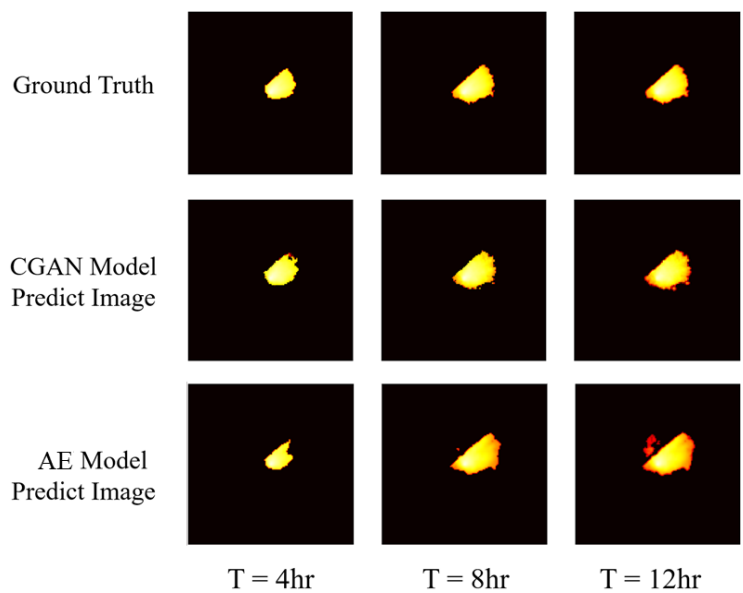

(b)

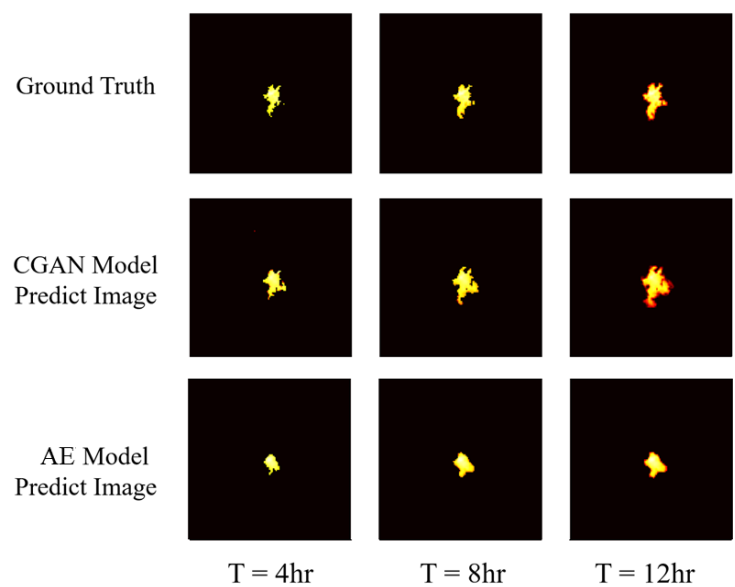

(c)

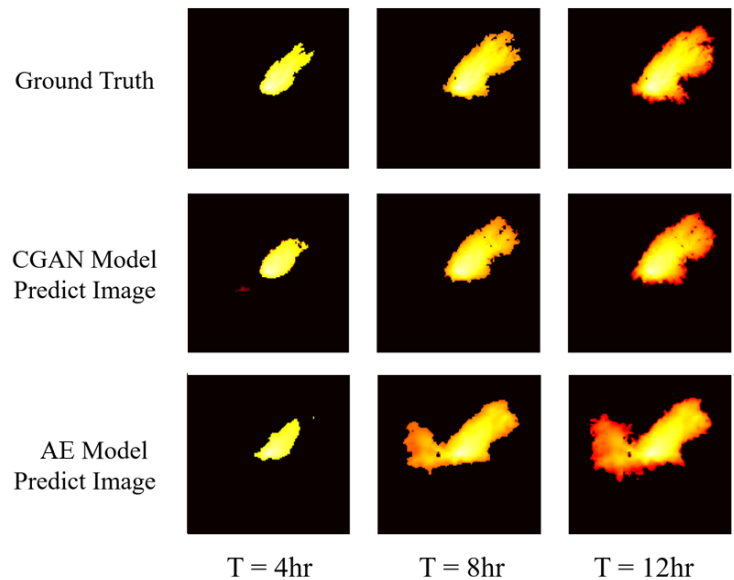

<table>
<tr><td>Ground Truth</td><td></td><td></td><td></td></tr>
<tr><td>CGAN Model Predict Image</td><td></td><td></td><td></td></tr>
<tr><td>AE Model Predict Image</td><td></td><td></td><td></td></tr>
<tr><td></td><td>T = 4hr</td><td>T = 8hr</td><td>T = 12hr</td></tr>
</table>

(d)

**Figure 14.** Comparison of image generation results from the CGAN-based model and the AE-based model at different time steps (4-, 8-, and 12-hours) for four representative test samples. The top row presents the ground-truth wildfire spread images, followed by predictions generated by the CGAN-based model and the AE-based model, respectively

## 4.3. Prediction for an Unseen Wildfire Scenario

Figure 15 presents a case study designed to evaluate the generalization capability of the proposed CGAN-based model. For this evaluation, a wildfire scenario is simulated in a different location (40.69186 °N, 120.460291 °W) that, while geographically similar to the training region, is not included in the construction of the training dataset. Furthermore, the simulation utilizes meteorological data from August 2023, a period not covered in the training data. Using this data, the FARSITE simulator is employed to generate ground-truth wildfire spread images, shown in the first row of Figure 15. Predictions from the CGAN-based and AE-based models are presented in the second and third rows, respectively.

The results on this scenario are consistent with the findings from the numerical investigations in Section 4.2. The CGAN-based model (second row) produces predictions that closely align with the ground-truth fire spread, successfully reproducing the irregular, complex, and highly detailed perimeter observe. By contrast, the AE-based model (third row) even fails to capture the overall trend of fire spread. Quantitative analysis is also performed in the same manner as the training and test datasets. For the 12-hour prediction, the MSE of the CGAN-based model (0.0876) is lower than that of the AE-based model (0.3809). However, the SSIM score of the AE-based model (0.8658) is higher than that of the CGAN-based model (0.8335), showing conflicting results. This discrepancy arises because most pixel intensities predicted by the AE-based model remain close to zero, which artificially inflates the SSIM scores despite the model's poor visual and structural correspondence with the ground truth. These findings confirm that the CGAN-based framework demonstrates strong generalization capability to previously unseen wildfire events. The model's robustness suggests that it can efficiently learn the governing mechanisms of wildfire propagation across diverse environmental and meteorological contexts, making it well-suited for future transfer learning applications in operational wildfire forecasting.

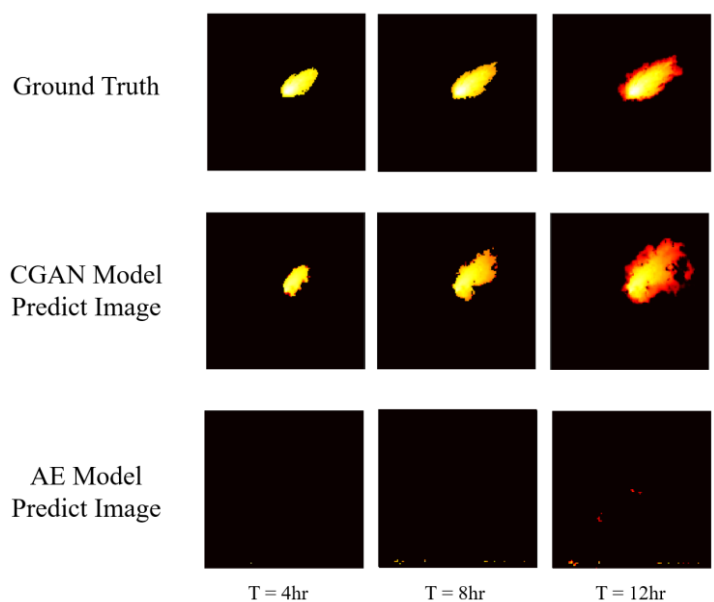

**Figure 15**. Comparison of image generation results for two different models (CGAN-based model and AE-based model) at different time points (4-, 8-, and 12-hour) and conditions for a new wildfire scenario. The top row shows the ground truth images, followed by the images predicted by the CGAN-based model and the AE-based model, respectively.

In addition to accuracy, computational efficiency is an important factor for practical use. For this case, the traditional physics-based simulator (FARSITE) requires 38 seconds to generate a prediction. By comparison, the AE-based model completes its prediction in 0.19 seconds, while the CGAN-based model requires 0.8 seconds. Note that the CGAN-based model appears slightly slower; however, the AE-based model produces a single prediction for the final time horizon (i.e., a total of three predictions for 4-, 8-, and 12-hour), whereas the CGAN-based model sequentially generates three intermediate time steps (4-, 8-, and 12-hour) and performs an ensemble of five simulations per step, effectively executing 15 predictions in total.

## 5. Conclusions

This study proposed a conditional generative adversarial network (CGAN)-based autoregressive model for probabilistic wildfire prediction, aiming to address the limitations of both traditional physics-based simulators and conventional deep learning models. The proposed model was trained on a large-scale dataset generated using the FARSITE simulator and evaluated against both an AE-based baseline model and FARSITE. The experimental results demonstrate the practical value of the proposed approach. The CGAN-based model achieves prediction accuracy comparable to FARSITE-generated wildfire spread image while operating approximately 47.5 times faster. This fast prediction capability is particularly crucial for supporting time-sensitive decision-making during the early stages of wildfire response. Compared to the AE-based model, the CGAN-based framework yields qualitatively superior predictions, producing sharper and more structurally faithful fire boundaries, while avoiding the over-smoothed spread patterns typically caused by pixel-wise loss functions. Its superior performance in reproducing the fire area during the intermediate stages of wildfire spread will significantly contribute to developing more effective response strategies for each timeframe during a wildfire outbreak. We also introduced a probabilistic framework for uncertainty quantification, enabling ensemble-based predictions that estimate the likelihood of fire occurrence across different locations.

While the proposed framework demonstrates superior predictive performance, the study acknowledges an important limitation: the model was trained exclusively on simulated data within a fixed domain size. This restricts its capacity to generalize to extremely large-scale or highly irregular wildfire events. Addressing this limitation is essential for practical deployment in real-world wildfire management. To this end, several research directions can be proposed.

- Considering historical wildfire observations: A crucial step forward is the integration of real-world wildfire observations (e.g., satellite or aerial imagery) into the training pipeline. Techniques such as domain adaptation, Bayesian updating, and transfer learning could be applied to bridge discrepancies between simulated and real data distributions.
- Enhancing spatiotemporal modeling: The predictive capacity of the model can also be strengthened by designing computationally efficient architectures capable of processing higher-resolution imagery. Current input resolution may limit the capture of fine-grained terrain or vegetation features that critically influence wildfire dynamics.
- Advancement of probabilistic prediction: Although ensemble-based stochastic sampling is employed in this study to quantify prediction uncertainty, a systematic sensitivity analysis regarding ensemble size and the selection of probabilistic confidence levels is necessary. Such an analysis would provide a deeper understanding of uncertainty propagation and improve the practical interpretability and reliability of the probabilistic forecasts.
- Integrating physics-informed approaches: To reduce the "black-box" nature of deep learning models, future work should consider the integration of simplified physical principles into the training process. Physics-informed machine learning can incorporate known fire dynamics, such as energy balance or fuel consumption laws, into the loss function as soft constraints, thereby guiding the model toward outputs that are not only data-consistent but also physically plausible. Furthermore, concepts from statistical physics, such as Percolation Theory, may provide a principled framework for modeling fire spread as a connectivity-driven process, where propagation occurs only when fuel density exceeds a critical threshold.
- Extending prediction horizons and fuel diversity: Another important direction is extending the model's predictive horizon beyond the current 12-hour time frame. Longer forecasts (e.g., 24–72 hours) would significantly improve planning and resource allocation for firefighting operations. Achieving this requires strategies to mitigate error accumulation over longer autoregressive sequences, potentially through hybrid methods that combine short-term neural forecasts with physics-based corrections or adopting more advanced deep learning techniques such as ConvLSTM or Transformer-based encoder. In addition, expanding the training dataset to include more diverse fuel types and ecosystems such as grasslands, chaparrals, or mixed-conifer forests would enhance the model's versatility across different wildfire regimes.

**Acknowledgment:** This study was carried out with the support of R&D Program for Forest Science Technology (Project No. RS-2025-25438293) provided by Korea Forest Service (Korea Forestry Promotion Institute).